\documentclass[runningheads]{llncs}
\usepackage{graphicx}

\usepackage{tikz}
\usepackage{comment}
\usepackage{amsmath,amssymb} 
\usepackage{color}
\usepackage{kotex}
\usepackage{multirow}
\usepackage{booktabs}
\usepackage[accsupp]{axessibility}  

\makeatother

\begin{document}
\pagestyle{headings}
\mainmatter
\def\ECCVSubNumber{100}  

\title{1st Place Solution to ECCV 2022 Challenge on HBHA: Transformer-based Global 3D Hand Pose Estimation in Two Hands Manipulating Objects Scenarios} 

\titlerunning{Transformer-based Global 3D Hand Pose Estimation} 


\author{Hoseong Cho$^{*}$ \and Donguk Kim$^{*}$ \and Chanwoo Kim \and\\ Seongyeong Lee \and Seungryul Baek}


\authorrunning{Cho et al}
%
\institute{Ulsan National Institute of Science and Technology (UNIST), South Korea\\
\email{\{hoseong,dukim,sky9739,skwithu,srbaek\}@unist.ac.kr}}


\maketitle

\begin{abstract}
    This report describes our 1st place solution to ECCV 2022 challenge on Human Body, Hands, and Activities (HBHA) from Egocentric and Multi-view Cameras (hand pose estimation). In this challenge, we aim to estimate global 3D hand poses from the input image where two hands and an object are interacting on the egocentric viewpoint. Our proposed method performs end-to-end multi-hand pose estimation via transformer architecture. In particular, our method robustly estimates hand poses in a scenario where two hands interact. Additionally, we propose an algorithm that considers hand scales to robustly estimate the absolute depth. The proposed algorithm works well even when the hand sizes are various for each person. Our method attains 14.4 mm (left) and 15.9 mm (right) errors for each hand in the test set.\footnotetext{These authors contributed equally to this work}
\end{abstract}

\section{Introduction}
\label{sec:intro}

The task of predicting hand poses that interact with an object is a core technique which could be widely applied to augmented reality (AR), virtual reality (VR), and robotics. Previous works have mainly dealt with single bare hands~\cite{baek2018augmented,baek2019pushing,kim2021end,moon2020interhand2} and object interaction scenarios~\cite{baek2020weakly,chao2021dexycb,garcia2018first,hampali2020honnotate,park2022handoccnet,sridhar2016real} with single hands. However, in the real-world, people use both hands when interacting with objects. The H2O dataset~\cite{kwon2021h2o} was recently proposed to research on this. This challenge is conducted on the H2O dataset that provides both hands' poses, object poses, and action classes. 

Existing hand pose estimation algorithms assumes that the tight bounding box of hands are already given by the detector. The pose estimation is mostly performed using the cropped images by the tight bounding boxes as~\cite{park2022handoccnet}; however this is not the practical setting when achieving the 3D pose estimation for two hands and an object, as the detection itself is frequently a hurdle due to the severe occlusions made by two hands and an object. To properly solve the problem, we propose a transformer-based global 3D hand pose estimation in hand-object interacting scenarios, which detects multiple hand joints at once in the entire image, to estimate the hand depth considering the entire context. In addition, we proposed the robust depth rescaling algorithm based on the hand scale. This can strongly predict the global pose estimation even when an image of the subject's hand, which was not seen during training, is used as input.  Details are given in the remainder of this document.

\begin{figure*}[!t]
\centering
\includegraphics[width=1\linewidth]{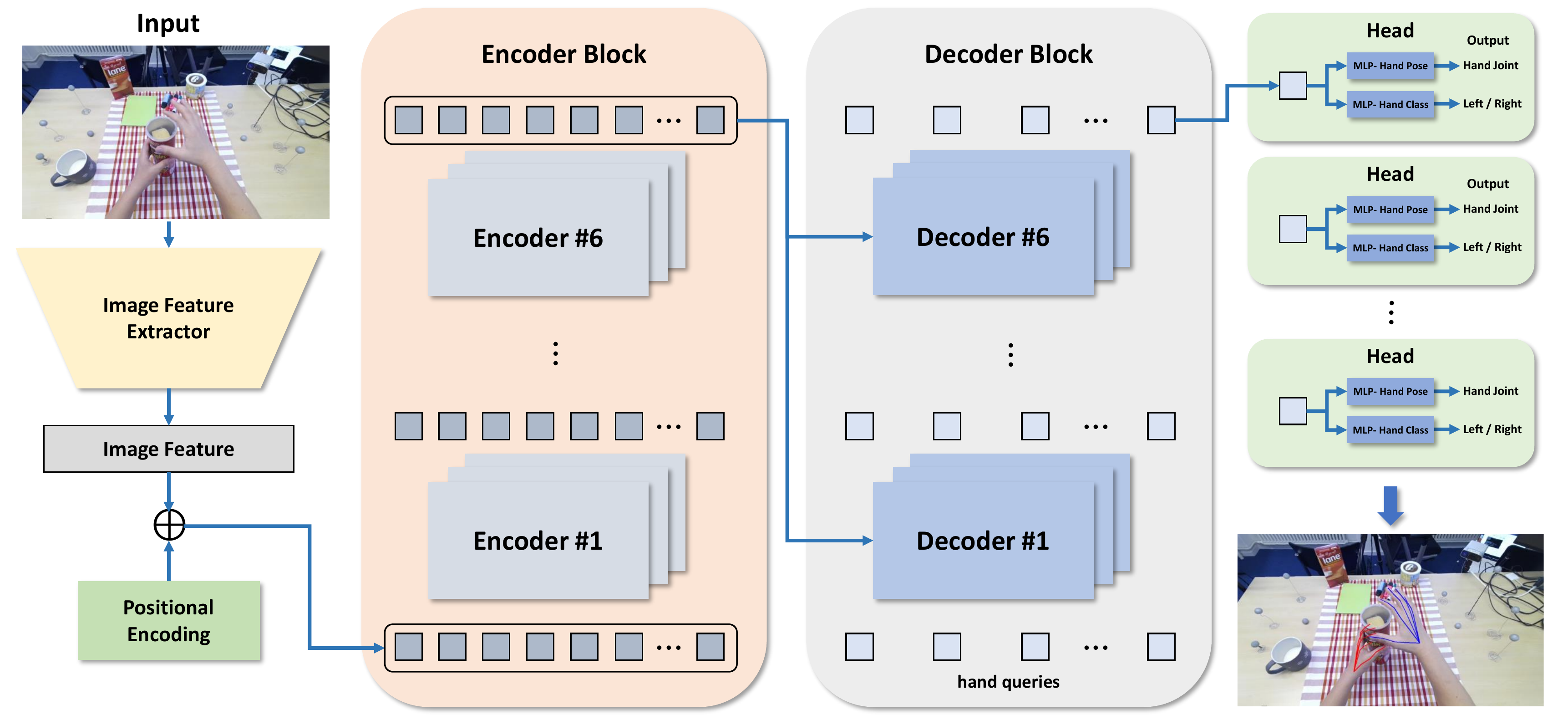}
\caption{Schematic diagram of our DETR-based~\cite{carion2020end} model. We use an egocentric-viewed images that include interactions between two hands and an object. Our model predicts x, y and z coordinates of two hands and their types (left/right hands) on each head.}
\label{fig:model}
\end{figure*}

\section{Methods}
\label{sec:intro}
In this section, we introduce our transformer-based architecture denoted in Figure~\ref{fig:model} and problem-solving mechanisms.

\subsection{Hand Pose Estimation based on Transformer}

DETR~\cite{carion2020end} is a transformer-based method that performs the object detection without much additional processes such as anchor generation and non-maximal supression (NMS). This model conducts the bipartite matching for (object class, bounding box) set prediction. To exploit the architecture towards estimating 3D hand joints, we modified their prediction heads. In detail, the head parts consist of two Multi-layer Perceptrons (MLPs) for predicting 2 hand types (ie. left, right hands) and 21 hand joints on UVD space. The output joints on UVD space can be converted to 3D x, y, z coordinates on the world space via camera intrinsic parameters afterwards. For training DETR model, the cross-entropy loss is involved to learn the hand type classification and the L1 loss to close the distance between the predicted 21 hand joints and their corresponding ground-truths.

\subsection{Hand Rescaling}
The problem of predicting the size and absolute depth of hands is ill-defined. This is due to the fact that RGB images do not have any clues for the depth information so that the network cannot properly predict depth values and actual scales of hands seeing solely on RGB images. Even for the same human's hands, the predicted absolute scales of left and right hands would become different, if two hands look in different sizes in the same images. 
Our network predicts the absolute depth and scale by looking at the whole image. After this, we use the mean scale of hands in train set to re-scale the absolute depth of two hands. Via the process, we could align the depth scales of both hands.

\section{Experiments}
\subsection{Experimental Settings}
\noindent \textbf{Implementation Details.}
The input image size is $1280 \times 720$. The random horizontal flip was applied as a data augmentation. We used AdamW optimizer with a different laerning rate: we used the learning rate of $10^{-4}$ for transformer and used the learning rate of $10^{-5}$ for the backbone network (ie. ResNet50). For each, we used the weight decay of $10^{-4}$. We used 4 RTX 3090 GPUs per each training and the batch size for each GPU was set as 4. The number of layers is set as $6$ for both transformer encoder and decoder. Training takes $300$ epochs with a learning rate drop by a factor of $10$ after $200$ epochs.

\noindent \textbf{Evaluation Metrics.}
Our results followed the mean hand pose error (left/right) metrics. In particular, unlike other metrics, the global 3D pose itself is compared without aligning the root depth of the GT hand with the position of the predicted hand.

\noindent \textbf{Evaluation Data.}
We use $3$ subject videos for training and evaluate performance with the last $1$ subject video for testing. Also, all data uses only egocentric views.

\setlength{\tabcolsep}{4pt}
\begin{table}
\begin{center}
\begin{tabular}{lll}
\hline\noalign{\smallskip}
Method & Left h. & Right h. \\
\noalign{\smallskip}
\hline
\noalign{\smallskip}
Hasson~\cite{hasson2020leveraging} & 39.5 & 41.8\\
H$+$O~\cite{tekin2019h+} & 41.4 & 38.8\\
H2O~\cite{kwon2021h2o} & 41.4 & 37.2\\ 
Ours & \textbf{14.4} & \textbf{15.9}\\
\hline
\end{tabular}
\end{center}
\caption{Pose errors (in mm.) of state-of-the-art methods on H2O. Best results are highlighted in \textbf{bold}.}
\label{table:quan}
\end{table}
\setlength{\tabcolsep}{1.4pt}

\subsection{Evaluation Results}
\begin{figure*}[!t]
\centering
\includegraphics[width=1\linewidth]{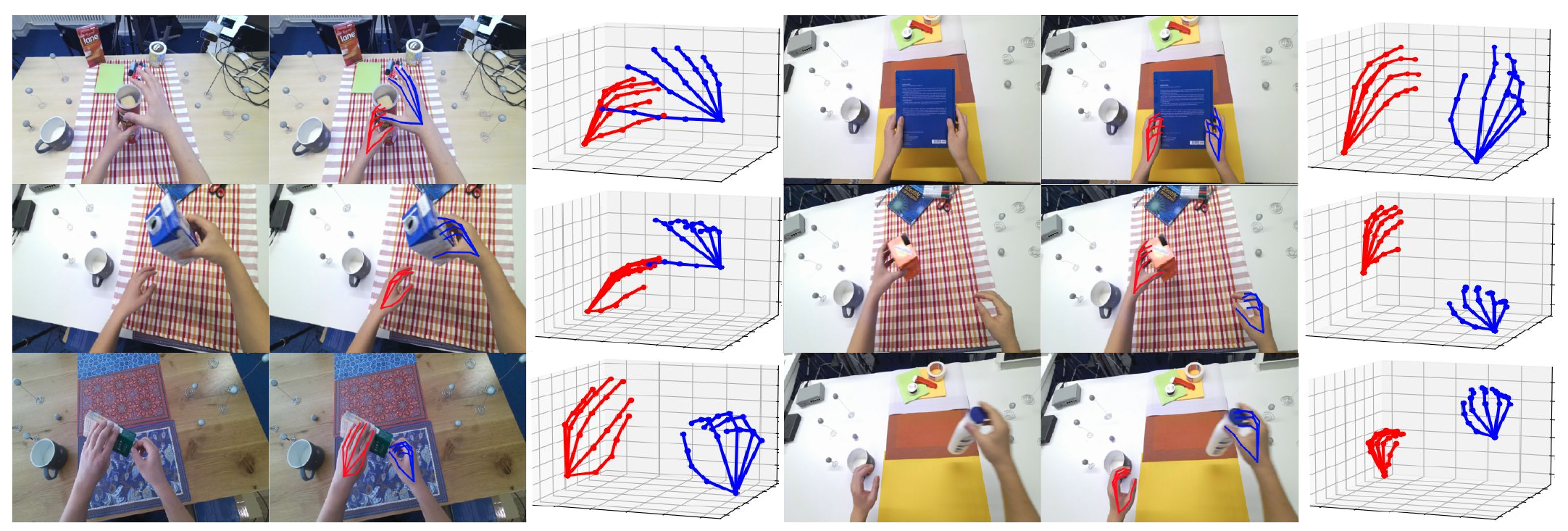}
\caption{Qualitative Results of our model on H2O dataset.}
\label{fig:vis}
\end{figure*}

We compared our model with state-of-the-art methods on H2O test set. Table~\ref{table:quan} shows that our proposed method significantly outperforms existing methods. Some example results are visualized in Figure~\ref{fig:vis}.


\begin{table}[!ht]
\centering
\begin{minipage}[t]{0.45\linewidth}
\renewcommand{\arraystretch}{1}
\resizebox{\textwidth}{!}{
\begin{tabular}{ccc}
\hline\noalign{\smallskip}
Resolution & Left h. & Right h. \\
\noalign{\smallskip}
\hline
\noalign{\smallskip}
$960 \times 540$ (relative) & 19.7 & 18.7\\
$960 \times 540$ (absolute) & 14.6 & 15.4\\ 
$1280 \times 800$ (absolute) & \textbf{14.2} & \textbf{15.2}\\
\hline
\end{tabular}}
\vspace{1em}
\caption{EPE (in mm.) comparison between different input image sizes and type of output depth.}
\label{table:ablation1}
\end{minipage}
\hspace{0.5em}
\begin{minipage}[t]{0.45\linewidth}
\renewcommand{\arraystretch}{1}
\resizebox{\textwidth}{!}{
\begin{tabular}{ccc}
\hline\noalign{\smallskip}
Rescaling & Left h. &  Right h. \\
\noalign{\smallskip}
\hline
\noalign{\smallskip}
w/o Rescaling & 26.8 & 28.0\\
w Rescaling (Ours) & \textbf{14.4} & \textbf{15.9}\\
\hline
\end{tabular}}
\vspace{1em}
\caption{Impact of Hand Rescaling algorithm on 3D hand pose estimation.}
\label{table:ablation2}
\end{minipage}
\end{table}
\vspace{-1cm}

\subsection{Ablation Study}
We tested our model on H2O validation set to find the optimal setting. Table~\ref{table:ablation1} shows that using high-resolution images as input performs better, which is similar to DETR. And predicting the absolute depth of each joint individually performs better than predicting the absolute depth of the root joint and the relative depth of other joints. Since the subjects of train set and validation set were the same, the experiment of hand re-scaling is added separately on Table~\ref{table:ablation1}.
The result is represented in Table~\ref{table:ablation2}.

\section{Conclusion}
We described our 1st place solution to ECCV 2022 challenge on Human Body, Hands, and Activities (HBHA) from Egocentric and Multi-view Cameras (hand Pose Estimation). To estimate two hand poses, we proposed a transformer-based global 3D hand pose estimator and a hand rescaling module. Our proposed method outperforms existing methods and ranks first in the HBHA challenge.

\clearpage
%
%
\bibliographystyle{splncs04}
\bibliography{main}
\end{document}